\documentclass[11pt,a4paper]{article}
\usepackage[hyperref]{acl2021}
\usepackage{times}
\usepackage{latexsym}

\usepackage{microtype}

\title{Claim Matching Beyond English to Scale Global Fact-Checking}
\author{
    Ashkan Kazemi\\
    University of Michigan and Meedan\\
    ashkank@umich.edu\\\And
    
    Kiran Garimella \\
    MIT \\ 
    garimell@mit.edu \\\AND
    
    Devin Gaffney \\
    Meedan \\ 
    devin@meedan.com \\\And
    
    Scott A.\ Hale \\
    University of Oxford and Meedan\\
    scott@meedan.com
    }
\date{}
\aclfinalcopy

\usepackage{booktabs} 
\usepackage{graphicx} 
\usepackage{mathtools} 
\usepackage{adjustbox} 
\usepackage{stfloats} 
\usepackage{siunitx} 
\begin{document}

\maketitle

\begin{abstract}
Manual fact-checking does not scale well to serve the needs of the internet. This issue is further compounded in non-English contexts. In this paper, we discuss claim matching as a possible solution to scale fact-checking. We define claim matching as the task of identifying pairs of textual messages containing claims that can be served with one fact-check. We construct a novel dataset of WhatsApp tipline and public group messages alongside fact-checked claims that are first annotated for containing ``claim-like statements'' and then matched with potentially similar items and annotated for claim matching. Our dataset contains content in high-resource (English, Hindi) and lower-resource (Bengali, Malayalam, Tamil) languages. We train our own embedding model using knowledge distillation and a high-quality ``teacher'' model in order to address the imbalance in embedding quality between the low- and high-resource languages in our dataset. We provide evaluations on the performance of our solution and compare with baselines and existing state-of-the-art multilingual embedding models, namely LASER and LaBSE. We demonstrate that our performance exceeds LASER and LaBSE in all settings. We release our annotated datasets\footnote{\url{https://doi.org/10.5281/zenodo.4890949}}, codebooks, and trained embedding model\footnote{\url{https://huggingface.co/meedan/indian-xlm-r}} to allow for further research.
\end{abstract}

\newlength{\extablecolwidth}
\setlength{\extablecolwidth}{0.85\columnwidth} 

\section{Introduction}
\begin{table*}
\begin{center}
\caption{Example message pairs in our data annotated for claim similarity.}
\label{table:example}
\adjustbox{max width=\textwidth}{\begin{tabular}{ p{\extablecolwidth} p{\extablecolwidth} c } 
\toprule
Item \#1 & Item \#2 & Label \\
\midrule
\raisebox{-0.75\totalheight}{\includegraphics[width=\extablecolwidth]{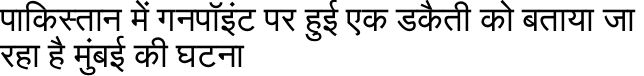}} & \raisebox{-0.75\totalheight}{\includegraphics[width=\extablecolwidth]{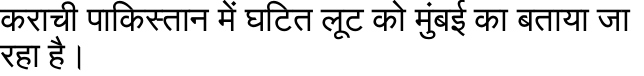}} & Very Similar\\
\midrule
\raisebox{-0.75\totalheight}{\includegraphics[width=\extablecolwidth]{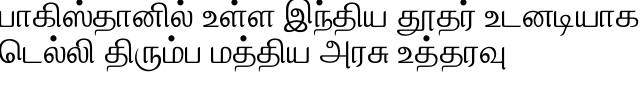}} & \raisebox{-0.75\totalheight}{\includegraphics[width=\extablecolwidth]{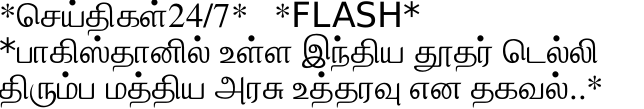}} & Very Similar\\
\midrule
Barber's salon poses the biggest risk factor for Corona! This threat is going to remain for a long duration. *At an average a barber's napkin touches 5 noses minimum* The US health dept chief J Anthony said that salons have been responsible for almost 50\% deaths. & *The biggest danger is from the barbershop itself*. This danger will remain for a long time. *Barber rubs the nose of at least 4 to 5 people with a towel,* The head of the US Department of Health J. Anthony has said that 50 percent of the deaths in the US have happened in the same way that came in saloons. & Very Similar \\
\midrule
\raisebox{-0.75\totalheight}{\includegraphics[width=\extablecolwidth]{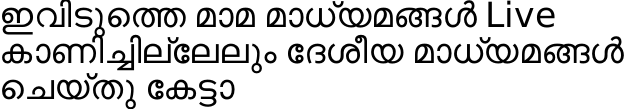}} & \raisebox{-0.75\totalheight}{\includegraphics[width=\extablecolwidth]{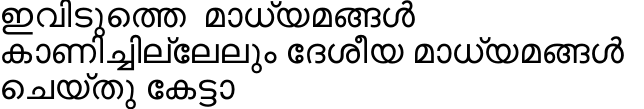}} & Very Similar\\
\midrule
Guys important msg:- There is the news of military bsf \& cisf coming to Mumbai and having a seven days to 2 weeks curfew.\ldots 
& *Just received information* Entire Mumbai and pune will be under Military lockdown for 10 days starts from Saturday.\ldots
& Somewhat Similar \\
\midrule
Don’t believe this FAKE picture of PM Modi; here’s the truth & Don’t believe this FAKE picture of Virat Kohli; here’s the fact check & Very Dissimilar \\
\bottomrule
\end{tabular}}
\end{center}
\end{table*}

Human fact-checking is high-quality but time-consuming. Given the effort that goes into fact-checking a piece of content, it is desirable that a fact-check be easily matched  with any content to which it applies. It is also necessary for fact-checkers to prioritize content for fact-checking since there is not enough time to fact-check everything. In practice, there are many factors that affect whether a message is `fact-check worthy' \cite{konstantinovskiy2020automated,hassan2017claimbuster}, but one important factor is prevalence. Fact-checkers often want to check claims that currently have high viewership and avoid fact-checking `fringe' claims as a fact-check could bring more attention to the claims---an understudied process known as amplification \cite{phillips2018oxygen, wardle2018lessons}. While the number of exact duplicates and shares of a message can be used as a proxy for popularity, discovering and grouping together multiple messages making the same claims in different ways can give a more accurate view of prevalence. Such algorithms are also important for serving relevant fact-checks via `misinformation tiplines' on WhatsApp and other platforms~\cite{wardle2019comprova, meedan2019checkpoint,magallon2019verificado}.

Identifying pairs of textual messages containing claims that can be served with one fact-check is a potential solution to these issues. The ability to group claim-matched textual content in different languages would enable fact-checking organizations around the globe to prioritize and scale up their efforts to combat misinformation. In this paper, we make the following contributions:
(i) we develop the task of claim matching, 
(ii) we train and release an Indian language XLM-R (I-XLM-R) sentence embedding model,
(iii) we develop a multilingual annotated dataset across high- and lower-resource languages for evaluation, and 
(iv) we evaluate the ability of state-of-the-art sentence embedding models to perform claim matching at scale. We formally evaluate our methods within language but also show clusters found using our multilingual embedding model often have messages in different languages presenting the same claims.

We release two annotated datasets and our codebooks to enable further research. The first dataset consists of 5,066 messages in English, Hindi, Bengali, Malayalam, and Tamil that have been triple annotated for containing `claim-like statements' following the definition proposed by fact-checkers in \citet{konstantinovskiy2020automated}. 
The second dataset consists of 2,343 pairs of social media messages and fact-checks in the same five languages as the first dataset annotated for claim similarity.
Table~\ref{table:example} shows examples of annotated pairs of messages from the second dataset.

\section{Related Work}
\subsection{Semantic Textual Similarity}
Semantic textual similarity (STS) refers to the task of measuring the similarity in meaning of sentences, and there have been widely adopted evaluation benchmarks including the Semantic Textual Similarity Benchmark (STS-B) \citeyearpar{cer-etal-2017-semeval, agirre-etal-2016-semeval, agirre-etal-2015-semeval, agirre-etal-2014-semeval, agirre-etal-2013-sem, agirre-etal-2012-semeval} and the Microsoft Research Paraphrase Corpus (MRPC) \cite{dolan-brockett-2005-automatically}. The STS-B benchmark assigns discrete similarity scores of 0 to 5 to pairs of sentences, with sentence pairs scored zero being completely dissimilar and pairs scored five being equivalent in meaning. The MRPC benchmark assigns binary labels that indicate whether sentence pairs are paraphrases or not. 

Semantic textual similarity is a problem still actively researched with a dynamic state of the art performance. In recent work from \citet{raffel2020exploring}, the authors achieved state-of-the-art performance on STS-B benchmark using the large 11B parameter T5 model. The ALBERT model \cite{lan2019albert} achieved an accuracy of 93.4\% on the MRPC benchmark and is considered one of the top contenders on the MRPC leaderboard.

While semantic textual similarity is similar to claim matching, the nuances in the latter require special attention. Claim matching is the task of matching messages with claims that can be served with the same fact-check and that does not always translate to message pairs having the same meanings. Moreover, claim matching requires working with content of variable length. In practice, content from social media also has wide variation in lexical and grammatical quality.

\subsection{Multilingual Embedding Models}
Embedding models are essential for claim and semantic similarity search at scale, since classification methods require a quadratic number of comparisons. While we have seen an increasing number of transformer-based contextual embedding models in recent years \cite{devlin-etal-2019-bert, reimers-gurevych-2019-sentence, cer-etal-2018-universal}, the progress has been asymmetric across languages.

The XLM-R model by \citet{conneau2019unsupervised} with 100 languages is a transformer-based model with a 250K token vocabulary trained by multilingual masked language modeling (MLM) with monolingual data and gained significant improvements in cross-lingual and multilingual benchmarks. LASER \cite{artetxe-schwenk-2019-massively} provided language-agnostic representation of text in 93 languages. The authors trained a BiLSTM architecture using parallel corpora and an objective function that maps similar sentences in the same vicinity in a high-dimensional space. Language-agnostic BERT sentence embeddings (LaBSE) by \citet{feng2020language} improved over LASER in higher resource languages by MLM and translation language modeling (TLM) pretraining, followed by fine-tuning on a translation ranking task \cite{yang2019improving}. 

\subsection{Claim Matching}
\citet{shaar-etal-2020-known} discussed retrieval and ranking of fact-checked claims for an input claim to detect previously debunked misinformation. They introduced the task, as well as a dataset covering US politics in English, and two BM25 based architectures with SBERT and a BERT-based reranker on top. \citet{vo-lee-2020-facts} tackled a similar problem by finding relevant fact-check reports for multimodal social media posts. However these projects only focus on English data that mainly cover U.S. politics and at least one of the matching pairs is a claim from a fact-check report. Additionally, the data collection process used in \citet{shaar-etal-2020-known} might not necessarily capture all possible matches for a claim, since the dataset is constructed by including only the claims mentioned in one fact-check report and not all previous occurrences. This may skew results and increase the risk of the model having a high false negative ratio. Recently, the CheckThat! Lab 2020 \cite{barron2020checkthat} has presented the same problem as a shared task. We improve on prior work by finding a solution that works for high- and low-resource languages and also for matching claims between pairs of social media content and pairs of fact-checks. We explicitly annotated claim pairs that might match, avoiding the aforementioned false negatives issue by design and providing more accurate models and evaluations.

\section{Data Sources}
The data used in this paper comes from a variety of sources. 
We use a mixture of social media (e.g., WhatsApp) content alongside fact-checked claims, since it is essential for any claim-matching solution to be able to match content both among fact-checked claims and social media posts as well as within social media posts. Among the prevalent topics in our data sources are the COVID-19 pandemic, elections, and politics.

\paragraph{Tiplines.}
Meedan, a technology non-profit, has been assisting fact-checking organizations to setup and run misinformation tiplines on WhatsApp using their open-source software, Check. 
A \textit{tipline} is a dedicated service to which `tips' can be submitted by users. On WhatsApp, tiplines are phone numbers to which WhatsApp users can forward potential misinformation to check for existing fact-checks or request a new fact-check.
The first tipline in our dataset ran during the 2019 Indian elections and received 37,823 unique text messages. Several additional always-on tiplines launched in December 2019 and ran throughout the 2020 calendar year. We obtained a list of the text of messages and the times at which they were submitted to these tiplines for March to May 2019 (Indian election tipline) and for February 2020 to August 2020 (all other tiplines). We have no information beyond the text of messages and the times at which they were submitted. In particular, we have no information about the submitting users.

\paragraph{WhatsApp Public Groups.}
In addition to the messages submitted to these tiplines, we have data from a large number ``public'' WhatsApp groups collected by \citet{garimella2020images} during the same time period as the Indian election tipline.
The dataset was collected by monitoring over 5,000 public WhatsApp groups discussing politics in India, totaling over 2 million unique posts. For more information on the dataset, please refer to \citet{garimella2020images}.
Such public WhatsApp groups, particularly those discussing politics have been shown to be widely used in India~\cite{lokniti2018}.

\paragraph{Fact-Check Reports.}
We aggregate roughly 150,000 fact-checks from a mixture of primary fact-checkers and fact-check aggregators. We employ aggregators such as Google Fact-check Explorer,\footnote{\url{https://toolbox.google.com/factcheck/explorer}} GESIS~\cite{tchechmedjiev2019claimskg}, and Data Commons, and include roughly a dozen fact-checking organizations certified by the International Fact-Checking Network with either global or geographically-relevant scope in our dataset. All fact-checks included at minimum a headline and a publish date, but typically also include a lead or the full text of the fact-check, as well as adjudication of the claim (e.g., truth or falsity), and sometimes include information of lesser value for our work such as author, categorization tags, or references to original content that necessitated the fact-check.

\section{Data Sampling \& Annotation}
To construct a dataset for claim matching, we design a two-step sampling and annotation process. We first sample a subset of items with potential matches from all sources and then annotate and select the ones containing ``claim-like statements.''

In a second task, we annotate pairs of messages for claim similarity. One of the messages in each pair must have been annotated as containing a ``claim-like statement'' in the first annotation task. We sample possible matches in several ways in order to not unnecessarily waste annotator time. We describe these sampling strategies and other details of the process in the remainder of this section.

\subsection{Task 1: Claim Detection}

Task 1 presented annotators with a WhatsApp message or fact-check headline and asked whether it contained a ``claim-like statement.'' 

We first created a codebook by inductively examining the English-language data, translations of the other-language data, and discussing the task with two fact-checkers (one Hindi-speaking and one Malayalam-speaking). We began with the definition set out by practitioners \cite{konstantinovskiy2020automated} for a ``claim-like statement'' and created examples drawn from our data sources. 
Annotators were asked whether the message had a claim-like statement and allowed to choose ``Yes'', ``Probably'', ``No'', or ``N/A: The message is not in language X'' (where X was the language being annotated). The instructions made clear ``Probably'' should be used sparingly and was intended for instances where an image, video, or other context was missing. The detailed instructions and an example of the interface are provided in the supplemental materials.

\begin{table}[tb]
    \centering
    \caption{Claim-like statements. $\kappa$ is Randolph's marginal-free kappa agreement on the collapsed data (Yes/Probably, No, Incorrect language). All languages were annotated by three annotators.}
    \begin{tabular}{lrrrr}
    \toprule
    Language & Items & $\kappa$ & Majority Yes\\
\midrule
Bengali (bn) & 1093 & 0.30 & 30\%\\ 
English (en) & 1000 & 0.60 & 54\%\\ 
Hindi (hi) & 1000 & 0.59 & 41\%\\ 
Malayalam (ml) & 1025 & 0.63 & 69\%\\ 
Tamil (ta) & 948 & 0.63 & 21\%\\ 
\bottomrule
    \end{tabular}
    
    \label{tbl:claimnoclaim}
\end{table}

We recruited three native speakers for each of Hindi, Bengali, Tamil, and Malayalam through Indian student societies at different universities as well as independent journalists. All of our annotators had a Bachelor's degree and many were pursuing Masters or PhDs. We onboarded all annotators and discussed the risks of possibly politically charged, hateful, violent, and/or offensive content in the dataset. Our custom-built annotation interface provided the ability to skip any piece of content with one keystroke. We also encouraged annotators to take frequent breaks and calculated these breaks into our payments. 

Our English-language data is a mix of Indian and global content. Two of our English annotators had previously completed the Hindi and Malayalam tasks while the third English annotator completed only the English-language task.

We calculate agreement using Randolph's marginal-free kappa \cite{randolph2005free}. This measure better estimates intercoder agreement in unbalanced datasets compared to fixed-marginal scores like Fleiss' kappa \cite{warrens2010inequalities}.

All participants annotated 100 items independently. We then discussed disagreements on these 100 items and updated the codebook if needed. The participants then annotated datasets of approximately 1,000 items in each language. Information about this final annotation dataset is presented in Table \ref{tbl:claimnoclaim}. Agreement between annotators for this task is lower than the next task but on par with annotation tasks for hate speech and other `hard tasks'~\cite{del2017hate,ousidhoum2019multilingual} suggesting determining whether a message has a claim-like statement is harder than determining the similarity of the statements (Task 2).

\begin{table}[tb]
    \centering
    \caption{Task 2 dataset. $\kappa$ is Randolph's marginal-free kappa \protect\cite{randolph2005free} agreement on the collapsed data (Very Similar, Not Very Similar, N/A). 
    ``V.\ Sim.'' is the percentage of cases where two or more annotators indicated the pairs were ``Very Similar.''}
    \adjustbox{max width=\columnwidth}{\begin{tabular}{lrrrr}
    \toprule
         Lang. & Pairs & $\kappa$ & Annotators & V.\ Sim.\\
    \midrule
        bn & 644 & 0.64--0.68 & 2--3 & 6\%\\ 
        en & 398 & 0.69 & 4 & 15\%\\
        hi & 399 & 0.90 & 3 & 21\%\\
        ml & 604 & 0.91 & 3 & 7\%\\
        ta & 298 & 0.85 & 2 & 11\%\\
    \bottomrule
    \end{tabular}}
    \label{tbl:claimsim}
\end{table}

\subsection{Task 2: Claim Similarity} 

The second task presented annotators with two messages and asked how similar the claim-like statements were in the messages. Annotators were given a four-point scale (``Very Similar'', ``Somewhat Similar'', ``Somewhat Dissimilar'', and ``Very Dissimilar''). We prepared a codebook with clear instructions for each response and examples in consultation with the two fact-checkers and discussed it with all annotators before annotation began. Annotators could also select ``N/A: One or more of the messages is not in language X or does not contain a claim-like statement'').

Our initial testing showed the largest source of disagreement was between ``Somewhat Dissimilar'' and ``Very Dissimilar.'' We added guidance to the codebook but did not dwell on this aspect as we planned to collapse these categories together. We prioritize our evaluations on ``Very Similar'' or ``Somewhat Similar'' statements.

Although our goal is claim matching, this task asked annotators about the similarity of claim-like statements as the annotators were not all fact-checkers. We found asking the annotators to speculate about whether some hypothetical fact-check could cover both statements was unhelpful. 
Our codebook is constructed such that ``Very Similar'' pairs of messages could be served by one fact-check while ``Somewhat Similar'' messages would partially be served by the same fact-check. A link to the codebook is in the supplemental materials. 

The same annotators from Task 1 completed Task 2 with a few exceptions. One Tamil annotator was unable to continue due to time restrictions, and one Bengali annotator only completed part of the annotations (we calculate agreement with and without this annotator in Table~\ref{tbl:claimsim}). We added a fourth English annotator in case there was another dropout but all English annotators completed. Table~\ref{tbl:claimsim} shows a breakdown of the dataset by language. In general, agreement on this task, even among the same annotators as Task 1, was much higher than Task 1 suggesting claim similarity is an easier task than claim detection. The largest point of disagreement was around the use of the N/A label: discussing this with annotators we found it was again the disagreement about whether certain messages had claims leading to the disagreement.

\begin{figure}
    \centering
    \includegraphics[width=\columnwidth]{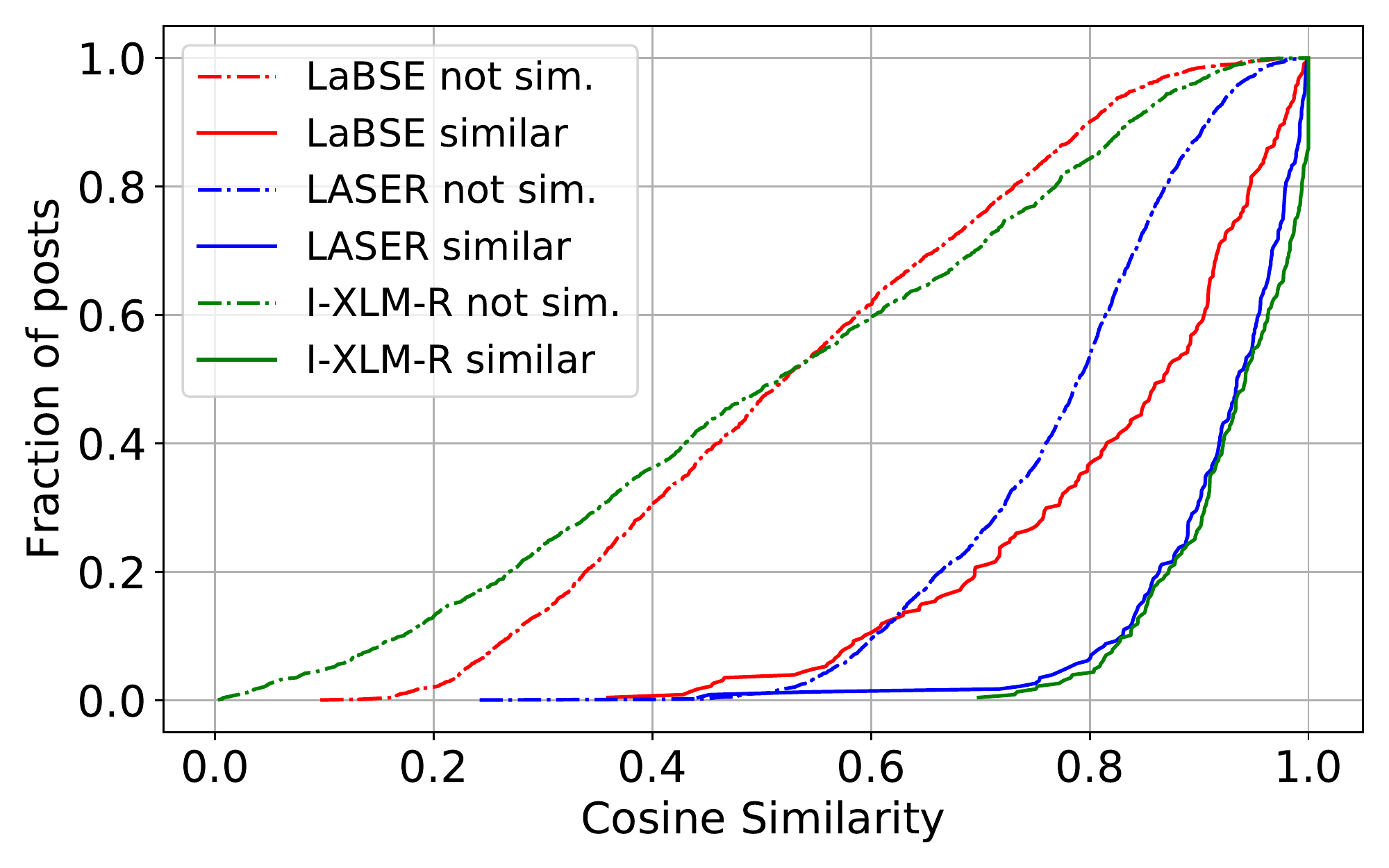}
    \caption{CDF of cosine similarities of all labeled data according to LASER, LaBSE, and I-XLM-R models. Legend: ``similar'' pairs were annotated by two or more annotators as being ``Very Similar''. ``not sim.'' encompasses all other pairs, excluding ``N/A'' pairs.}
    \label{fig:label_dist}
\end{figure} 

\subsection{Sampling}
A purely random sample of pairs is very unlikely to find many pairs that match. We considered examining pairs with the highest cosine similarities only, but these pairs were likely to match in trivial and uninteresting ways. In the end, we used random stratified sampling to select pairs for annotation.

We first calculate all pairwise cosine similarities using multiple embedding models (described in Section~\ref{sec:claim-matching-methods}). We then use stratified sampling to sample 100 pairs in proportion to a Gaussian distribution with mean 0.825 and standard deviation 0.1 for each model and language. We do this due to our strong prior that pairs close to zero as well as pairs close to one are usually `uninteresting.' These represent pairs that either clearly do not match or (very often) clearly match.
In practice, we still sample a wide range of values (Figure~\ref{fig:label_dist}). We also include 100 random pairs for each language with the exception of Tamil due to annotator time limitations.

We used LASER, LaBSE, and our Indian XLM-R (I-XLM-R) model (details below) to sample pairs for all languages. 
Our Bengali and Malayalam annotators had additional capacity and annotated additional pairs drawn in a similar way.

\section{Claim Matching Methods}\label{sec:claim-matching-methods}
\subsection{Experimental Setup}
We use a GPU-enabled server with one 1080 GPU to train our own embedding model and run the rest of our experiments on desktop computers with minimal runtime. We use the Elasticsearch implementation of the BM25 system and use the Sentence-Transformers (for I-XLM-R), PyTorch (for LASER), and TensorFlow (for LaBSE)\footnote{We use \url{https://github.com/bojone/labse}.} to train and retrieve embeddings. We follow the approach of \citet{reimers-gurevych-2020-making} for  tuning the hyperparameters of our embedding model.

\subsection{Training a Multilingual Embedding Model}\label{sec:msbert}
We use the knowledge distillation approach presented in \citet{reimers-gurevych-2020-making} to train a multilingual embedding model.\footnote{Trained models from \citeauthor{reimers-gurevych-2020-making} do not include embeddings for Bengali, Tamil, and Malayalam, which motivated us to train the I-XLM-R model.} The approach adopts a student--teacher model in which a high quality teacher embedding model is used to align text representations of a student model by mapping embeddings of text in the student language to close proximity of the embeddings of the same text in the teacher language. Using this approach we train a model for English, Hindi, Malayalam, Tamil, and Bengali. We refer to this model as our Indian XLM-R model (I-XLM-R), and use it as one of the models we evaluate for claim matching.

\paragraph{Training Data.}
The knowledge distillation approach requires parallel text in both student and teacher languages for training embedding models. We find the OPUS parallel corpora \cite{tiedemann-2012-parallel} to be a useful and diverse resource for parallel data. 
We retrieve parallel data between English and the collection of our four Indian languages from OPUS and use it as training data.

\paragraph{Training Procedure.}
For a teacher model $M_T$ and a student model $M_S$ and a collection of $(s_i, t_i)$ pairs of parallel text, we minimize the following MSE loss function for a given mini-batch B:

\noindent
\begin{center}
\small
\begin{math}
\frac{1}{|B|} \sum\limits_{i \in B}{ [( M_T(s_i) - M_S(s_i) )^2 + ( M_T(s_i) - M_S(t_i) )^2] }
\end{math}
\end{center}

Intuitively, this loss function forces embeddings of the student model for both $t_i$ and $s_i$ to be in proximity of the teacher embeddings for $s_i$, therefore transferring embedding knowledge from the teacher to the student model. For training our Indian XLM-R model, we pick the English SBERT model as teacher \cite{reimers-gurevych-2019-sentence} (for its high quality embeddings) and XLM-Roberta (XLM-R) as the student (for SOTA performance in NLP tasks and a universal vocabulary that includes tokens from 100 languages).

\subsection{Model Architecture}
We evaluate a retrieval-based claim matching solution built on top of the BM25 retrieval system \cite{robertson2009probabilistic} as well as an embeddings-only approach. In the first case, queries are fed into BM25 and the retrieved results are then sorted based on their embedding similarity to the input query. The top ranking results are then used as potential matches for the input claim.
In the latter case, we classify pairs of items using features derived from the embedding models. 

\section{Results}\label{sec:results}
For some applications, it is good enough to be able to rank the most similar claims and treat the problem of claim matching as an information retrieval problem. This is the case, for example, when fact-checkers are examining possible matches to determine if a new content item matches a previous fact-check. We discuss the performance of information retrieval approaches in Section~\ref{sec:results-ir}.

In many other applications, however, we seek a system that can determine if the claims in two items match without human intervention. These applications demand a classification approach: i.e., to determine whether two items match.
This allows similar items to be grouped and fact-checkers to identify the largest groups of items with claims that have not been fact-checked. We discuss the performance of simple classification approaches in Section~\ref{sec:results-classification}.

\subsection{Information Retrieval Approach}
\label{sec:results-ir}
We find the mean reciprocal rank (MRR) metric to be a good IR-based performance measure for our system, since we only know of one match in the retrieved results by the system for our queries. We use the base BM25 system as a strong baseline to compare against. We also compare our system with other state-of-the-art multilingual embedding models used for reranking, namely LASER and LaBSE. Results are presented in Table~\ref{tab:mrr}.

\begin{table}[tb]
    \centering
    \caption{MRR across different models and languages. Columns refer to reranking embedding models on top of BM25, with the exception of BM25 as the baseline.}
    \adjustbox{max width=\columnwidth}{\begin{tabular}{lcccc}
    \toprule
         Language & BM25 & LASER & LaBSE & I-XLM-R \\
         \midrule
         Bengali & 0.4247 & 0.4170 & 0.4120 & \textbf{0.5281} \\
         English & \textbf{0.4286} & 0.4247 & 0.4101 & 0.4221 \\
         Hindi & 0.4524 & 0.4289 & 0.3675 & \textbf{0.4849} \\
         Malayalam & 0.3903 & 0.3777 & 0.3651 & \textbf{0.4023} \\
         Tamil & \textbf{0.4747} & 0.4050 & 0.4563 & 0.4634 \\
    \bottomrule
    \end{tabular}}
    
    \label{tab:mrr}
\end{table}

The BM25 with I-XLM-R reranking outperforms other systems in all languages, with the exception of Tamil and English where the system performs comparably with the BM25 baseline. The largest lead in performance of the I-XLM-R based model is for Bengali, where the MRR score is more than 0.1 higher than the BM25 baseline.

Both LASER and LaBSE fall short on surpassing the baseline for any of the languages. LASER performs the worst on Tamil, where its MRR score is nearly 0.07 less than BM25. Similarly, LaBSE's largest difference with BM25 is in Hindi where it falls short by 0.085. Although there is room for improvement in some languages, the I-XLM-R seems the best choice if only one system is chosen.

After calculating MRR we also evaluated the systems on other metrics, namely ``Mean First Relevant'' (MFR, \citet{fuhr2018some}) and HasPositive@K \cite{shaar-etal-2020-known}. 
Both measures did not demonstrate any meaningful patterns useful for selecting the best system.
We do not include the details of these evaluations for brevity.

\begin{figure}
    \centering
    \includegraphics[width=\columnwidth]{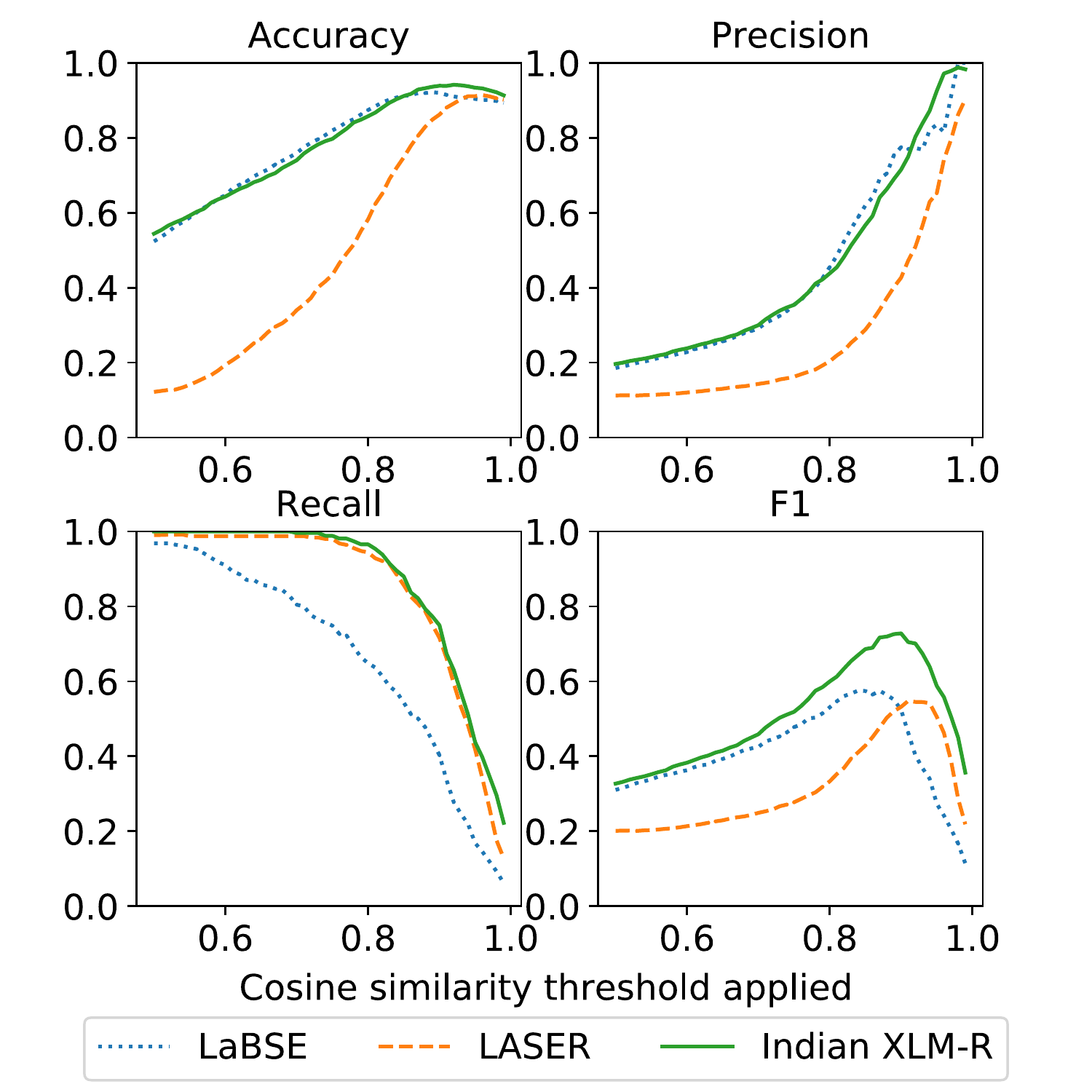}
    \caption{Accuracy, Precision, Recall, and F1 for simple thresholds on the cosine similarity scores.}
    \label{fig:thresholds}
\vspace{-\baselineskip}
\end{figure}

\subsection{Classification Approaches}
\label{sec:results-classification}
Responding to submitted content on a tipline, as well as grouping claims to understand their relative prevalence/popularity, requires more than presenting a ranked list as occurs in the information retrieval approaches in the previous subsection and in previous formulations of this problem \cite[e.g.,][]{shaar-etal-2020-known}. In this section we use the annotated pairs to evaluate how well simple classifiers perform with each model.

\paragraph{Threshold Classifier.} The first `classifier' we evaluate is a simple threshold applied to the cosine similarity of a pair of items. Items above the threshold are predicted to match while items with a similarity below the threshold are predicted to not match. In doing this, we seek to understand the extent to which the embedding models can separate messages with matching claims from those with non-matching claims. 

An ideal model would assign higher cosine similarity scores to every pair of messages with matching claims than to pairs of messages with non-matching claims. Table~\ref{tbl:threshold-f1-10fold} shows the F1 scores averaged across 10 runs of 10-fold cross validation for binary classifiers applied to all languages and each language individually. In general, the Indian XLM-R model performs best at the task with F1 scores 
ranging from 0.57 to 0.88. 
As shown in Figure~\ref{fig:thresholds}, our Indian XLM-R model outperforms LASER primarily in precision and outperforms LaBSE primarily in terms of recall.

\begin{table}[tb]
    \centering
    \caption{Maximum average F1 scores $\pm$ standard deviations achieved with 10 runs of 10-fold cross-validation and the corresponding thresholds (thres.) for each score. The `classifiers' are simple thresholds on the cosine similarities.}
    \adjustbox{max width=\columnwidth}{\begin{tabular}{lrrr}
    \toprule
    & \multicolumn{1}{c}{LASER} & \multicolumn{1}{c}{LaBSE} & \multicolumn{1}{c}{I-XLM-R}\\
    \cmidrule(lr){2-2}\cmidrule(lr){3-3}\cmidrule(lr){4-4}
    Language & F1 (thres.) & F1 (thres.) & F1 (thres.)\\
    \midrule
        All & 0.55$\pm$0.08& 0.58$\pm$0.07& \textbf{0.73}$\pm$0.07\\
             & (0.91)& (0.84)& (0.90)\\
        Bengali & 0.68$\pm$0.21& 0.58$\pm$0.23& \textbf{0.74}$\pm$0.19\\
             & (0.96)& (0.90)& (0.96)\\
        English & 0.85$\pm$0.09& 0.77$\pm$0.15& \textbf{0.88}$\pm$0.10\\
             & (0.85)& (0.77)& (0.78)\\
        Hindi & 0.74$\pm$0.13& 0.61$\pm$0.15& \textbf{0.82}$\pm$0.12\\
             & (0.88)& (0.87)& (0.87)\\
        Malayalam & 0.47$\pm$0.20& 0.71$\pm$0.20& \textbf{0.79}$\pm$0.20\\
             & (0.92)& (0.85)& (0.89)\\
        Tamil & 0.26$\pm$0.21& 0.50$\pm$0.20& \textbf{0.57}$\pm$0.15\\
             & (0.99)& (0.98)& (0.96)\\
    \bottomrule
    \end{tabular}}
    \label{tbl:threshold-f1-10fold}
\end{table}

The numbers reported in Table~\ref{tbl:threshold-f1-10fold}'s last column all come from I-XLM-R. The English-only SBERT model performs slightly better with a maximum F1 score of 
0.90$\pm$0.09 at a threshold of 0.71 
on English data, suggesting that the student model may have drifted from the teacher model for English during training. This drift is slight, however, and the cosine similarities across all English-language data for the two models are highly correlated with a Pearson's correlation coefficient of 0.93. The authors of SBERT released two additional multilingual models on that support English and Hindi, but do not support Bengali, Malayalam, or Tamil.\footnote{\url{https://www.sbert.net/docs/pretrained_models.html} has `xlm-r-distilroberta-base-paraphrase-v1' and `xlm-r-bert-base-nli-stsb-mean-tokens'} We find the models have comparable performance to I-XLM-R on English \& Hindi while F1 scores for other languages are between 0.17 and 0.61. 

Our dataset includes both social media messages (namely, WhatsApp messages) and fact-checks. Overall, performance is higher for matching fact-checks to one another than for matching social media messages to one another for all models. As an example, the best-performing model, Indian XLM-R, achieves a maximum F1 score of 0.76 with a threshold 0.87 for matching pairs of fact-checks, but only a maximum F1 score of 0.72 (threshold 0.90) for matching pairs of social media messages. 

\paragraph{Claim Matching Classifier.}
We train an AdaBoost binary classifier that predicts if two textual claims match. The features are all precomputed or trivial to compute so that such a system could easily be run to refine a smaller number of candidate matches with minimal additional computation.

We use lengths of claims, the difference in lengths, embedding vectors of each item, and their cosine similarity as features. We build a balanced dataset by taking all the ``Very Similar'' pairs and matching every item with a randomly selected ``Not Very Similar'' (every other label) item from the same language. We do not differentiate between pairs in different languages as our per language data is limited and all features including the embedding vectors translate across languages as they are from mulitilingual embedding models.

\begin{table}[tb]
    \centering
    \caption{Claim matching classification results.}
    \adjustbox{max width=\columnwidth}{\begin{tabular}{lccc}
    \toprule
         Model & Accuracy & F1 (+) & F1 (-) \\
         \midrule
         LASER & 0.805$\pm$0.064 & 0.789$\pm$0.087 & 0.814$\pm$0.039 \\
         LaBSE & 0.797$\pm$0.059 & 0.791$\pm$0.067 & 0.800$\pm$0.055 \\
         I-XLM-R & \textbf{0.883}$\pm$0.036 & \textbf{0.885}$\pm$0.036 & \textbf{0.880}$\pm$0.037 \\
         \midrule
         All & 0.868$\pm$0.036 & 0.868$\pm$0.036 & 0.866$\pm$0.039 \\
    \bottomrule
    \end{tabular}}
    \label{tab:adaboost}
\vspace{-\baselineskip}
\end{table}

Claim matching classification results are presented in Table~\ref{tab:adaboost}. We evaluate models using 10-fold cross validation and report accuracy and F1 scores for each class averaged over 10 runs. Consistent with previous outcomes, it is clear that using the I-XLM-R cosine similarity and embeddings as input features results in better performance than other models, including the model with all features.

The positive class F1 scores for all models in Table~\ref{tab:adaboost} are notably higher than the threshold approaches (Table~\ref{tbl:threshold-f1-10fold}) suggesting information from the embeddings themselves and the lengths of the texts are useful in determining whether the claims in two messages match.  The claim matching classifier is language-agnostic and is learning from only 522 datapoints, which underscores the quality of the I-XLM-R embeddings.

\paragraph{Error Analysis.}
We manually inspect the pairs classified in error using the ``threshold classifier'' and I-XLM-R. The pairs either have a similarity score above the matching threshold but are ``Not Similar'' (false positives, 24/89) or are matches and have a score below threshold (false negatives, 65/89). 16 of the 24 false positives are labeled as ``Somewhat Similar,'' and manual inspection shows that these pairs all have overlapping claims (i.e., they share some claims but not others). There are no obvious patterns for the false negatives, but some of the errors are made in ambiguous cases.

We also examine the errors of one random fold of the AdaBoost classifier to further investigate where our model makes mistakes. There are a total of 10 wrong predictions (6 false negatives and 4 false positives). Of these, 2/6 and 1/4 are annotation errors. Within the false negatives, most other cases are pairs of text that are very similar but minimally ambiguous because of a lack of context, which annotators correctly resolved to being identical. 
An example of such a false negative is the pair of messages ``Claim rare flower that blooms once in 400 years in the-himalayas-called-mahameru-pushpam'' and ``Images of Mahameru flower blooms once every 400 years in Himalayas.''
False positives were all ``Somewhat Similar'' and ``Somewhat Dissimilar'' pairs that the classifier mistook for ``Very Similar.'' There were no significant discrepancies among languages in classification errors.


\section{Discussion \& Conclusions} 

Scaling human-led fact-checking efforts requires matching messages with the same claims. In this paper, we train a new model and create an evaluation dataset that moves beyond English and American politics. Our system is being used in practice to support fact-checking organizations.

We find that the embedding models can generally match messages with the same claims. Performance for matching fact-checks slightly exceeds that for matching social media items. This makes sense, given that fact-checks are written by professional journalists and generally exhibit less orthographical variation than social media items.

Too few examples of fact-checks correctly matched a social media item to evaluate performance in that setting. This is not a major limitation since nearly every fact-check starts from a social media item. So, in practice we only need to be able to match social media items to one another in order to locate other social media items having the same claims as the item that led to a fact-check. %

We evaluate claim matching within each language, but the embedding models are all multilingual and could serve to match claims across languages.
BM25 is not multilingual, but Elasticsearch can index embeddings directly. Previously \citet{almeida2020text} developed a Elasticsearch plugin to query embeddings by cosine distance, but since version 7.3 of Elasticsearch this functionality is now available natively in Elasticsearch \cite{elasticsearchdocs2019}, meaning a large set of embeddings can be searched efficiently to find near matches across languages.

As a proof of concept, we took the 37,823 unique text messages sent to the Indian election tipline and clustered them using I-XLM-R and online, single-link hierarchical clustering with a threshold of 0.90. We found 1,305 clusters with 2 or more items; the largest cluster had 213 items.
We hired an Indian journalist with experience fact-checking during the Indian 2019 elections to annotate each of the 559 clusters with five or more items by hand.
The annotation interface presented three examples from each cluster: one with the lowest average distance to all other messages in the cluster, one with the highest distance, and one message chosen randomly. In 137 cases the examples shown for annotation were from multiple languages, and in 
132 of those cases
the journalist was able to identify the same claims across multiple languages. Although preliminary, this demonstrates the feasibility and importance of multilingual claim matching with these methods---an area we hope further work will tackle.

Our findings are supporting over 12 fact-checking organizations running misinformation tiplines. The deployed system uses I-XLM-R and automatically groups text messages with similarities over 0.95 and recommends possible matches from less-similar candidates that fact-checking organizations can confirm or reject. Matches can also be added manually. Initial feedback from the fact-checkers has been positive, and we are collecting data for further research and evaluation.

We prioritized the well-being of annotators and the privacy of WhatsApp users throughout this research. Our data release conforms to the FAIR principles~\cite{wilkinson2016fair}. We have no identifying information about WhatsApp users and any references to personally identifiable information in messages such as phone numbers, emails, addresses and license plate numbers are removed to preserve user privacy. We worked closely with our annotators preparing them for the risk of hateful content, encouraging frequent breaks, and paying well-above minimum wage. We took a compassionate response to COVID disruptions and other life stresses even when this meant less annotated data than was originally envisioned.


\section*{Acknowledgments}
This work was funded by the Omidyar Network with additional support from Sida, the Robert Wood Johnson Foundation, and the Volkswagen Foundation. Kiran Garimella is supported by the Michael Hammer postdoctoral fellowship at MIT. We are thankful to all of the wonderful annotators and fact-checking organizations who made this research possible. We are grateful to the Meedan team, Prof.\ Rada Mihalcea, Gautam Kishore Shahi, and our anonymous reviewers.

\bibliography{anthology,custom}
\bibliographystyle{acl_natbib}

\clearpage
\section{Supplemental Materials}
\subsection{Codebooks}

Our codebooks are available openly. Due to the page limit for the supplemental materials, we provide hyperlinks to these codebooks:
\begin{itemize}
    \item \href{https://docs.google.com/document/d/1yyo_h-52-w_S7ObHfyvDFUVabIJ09624cku5IHTIEFQ/edit?usp=sharing}{Claim detection codebook}
    \item \href{https://docs.google.com/document/d/1D88VAfTGL1vTVuMKhkX7AqDFuem3xkBpu6SFRsmy-3Q/edit?usp=sharing}{Claim similarity codebook}
\end{itemize}

\begin{figure*}
    \centering
    \includegraphics[width=\textwidth]{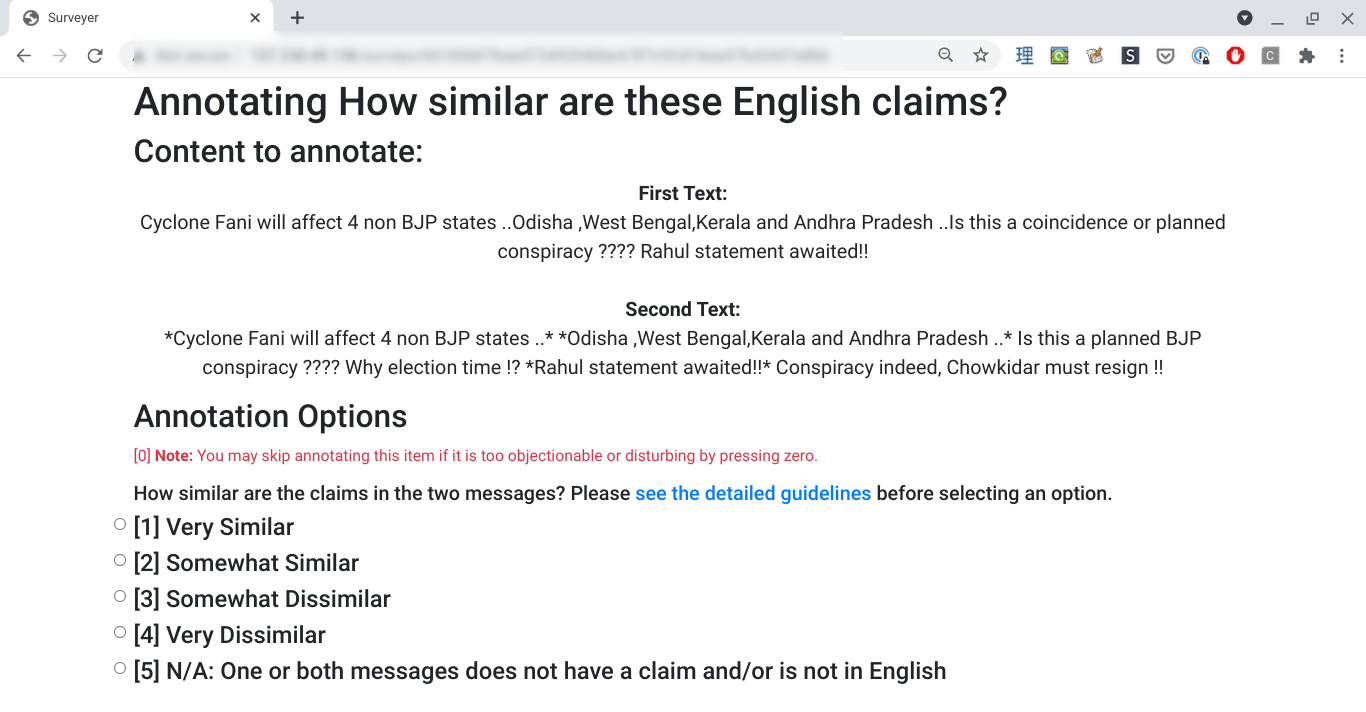}
    \caption{An example of the annotation interface}
    \label{fig:annotation-screen-capture}
\end{figure*}

We coded a simple annotation interface, which is free and open-source: \url{https://github.com/meedan/surveyer/}. A screen capture of the annotation interface during the English-language claim-similarity task is shown in Figure~\ref{fig:annotation-screen-capture}

\subsection{Per language results}
\begin{figure*}
    \centering
    \includegraphics[width=\textwidth]{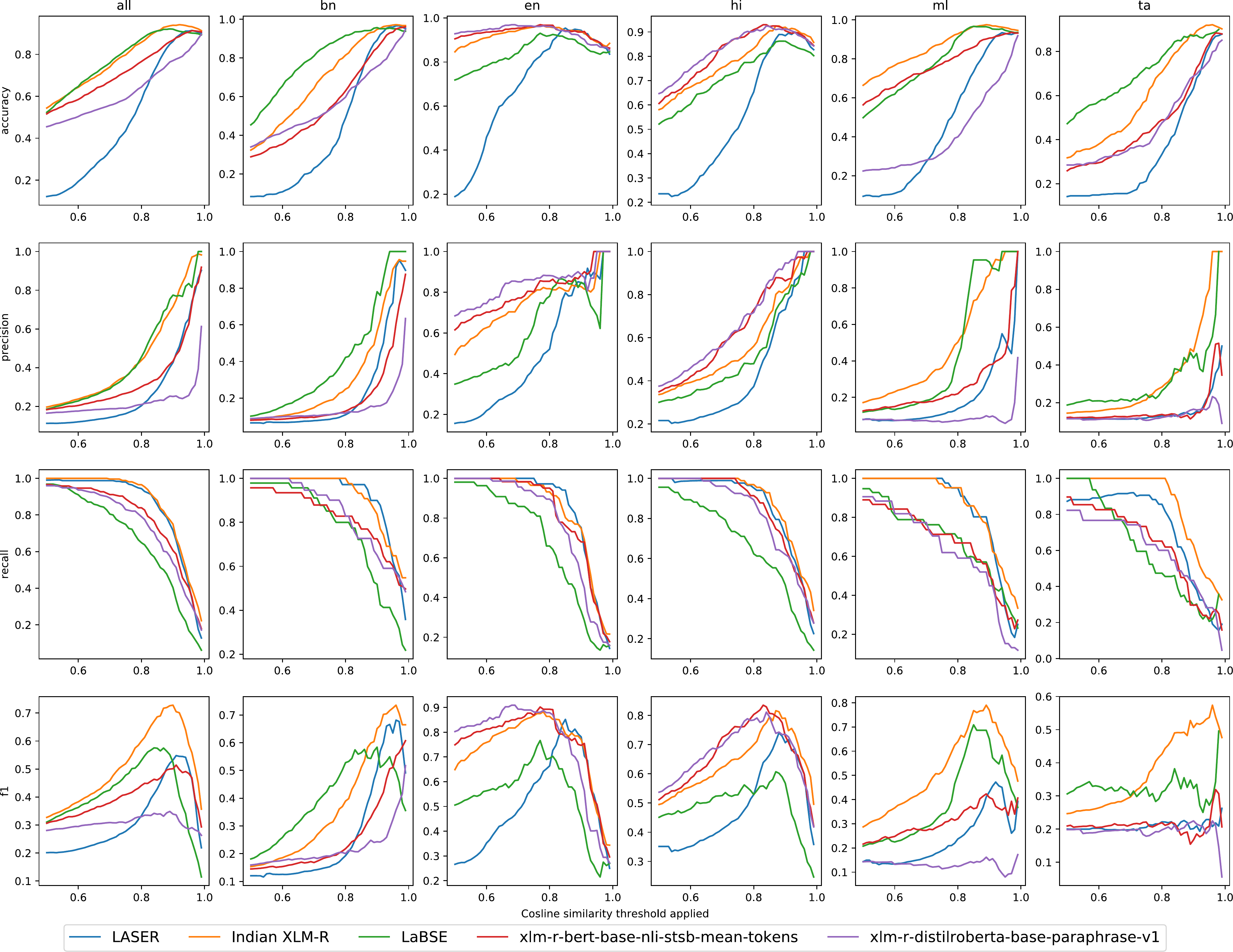}
    \caption{Accuracy, precision, recall, and F1 scores for each language individually. Positive class is ``Very similar.''}
    \label{fig:thresholdsperlang}
\end{figure*}

Figure~\ref{fig:thresholdsperlang} shows the accuracy, precision, recall, and F1 scores for simple threshold classifiers. This is equivalent to Figure~\ref{fig:thresholds}, but shows the plots for each language individually in addition to the overall values across all languages.

The figure also includes two additional embedding models from the SBERT website: \emph{xlm-r-distilroberta-base-paraphrase-v1} and \emph{xlm-r-bert-base-nli-stsb-mean-tokens}.\footnote{\url{https://www.sbert.net/docs/pretrained_models.html\#multi-lingual-models}} As discussed in the main paper, we find our models far outperform these models for Bengali, Malayalam, and Tamil while performance for English and Hindi is similar.


\subsection{Alternative definition of the positive class}

The analysis in the paper presents results for ``Very Similar'' compared to all other classes (N/A labels excluded). Here we show qualitatively similar results are obtained when the positive class is items for which a majority of annotators indicated ``Very Similar'' or ``Somewhat Similar.'' As stated, somewhat similar matches are useful as a fact-check would partially address some of the claims in a somewhat similar match. Table~\ref{tbl:claim-matching-distribution} provides the distribution of labels for the claim matching dataset.

Table~\ref{tbl:threshold-somewhatsim} presents F1 scores averaged across 10 runs of 10-fold cross validation using ``Somewhat Similar'' or ``Very Similar'' as the positive class. The results are similar to  Table~\ref{tbl:threshold-f1-10fold} in the main paper. F1 scores are generally higher, but our Indian XLM-R model still performs best. Surprisingly, LASER matches its performance in one language (Hindi).

\begin{table*}[h!]
    \centering
    \caption{Maximum F1 scores (F1) and standard deviations achieved and the corresponding thresholds (thres.) for each score. The `classifiers' are simple thresholds on the cosine similarities. Scores are the average of 10 rounds of 10-fold cross validation. \textbf{The positive class is ``Somewhat Similar'' or ``Very Similar.''}}
    \label{tbl:threshold-somewhatsim}
    \begin{tabular}{lrrr}
    \toprule
    & \multicolumn{1}{c}{LASER} & \multicolumn{1}{c}{LaBSE} & \multicolumn{1}{c}{I-XLM-R}\\
    \cmidrule(lr){2-2}\cmidrule(lr){3-3}\cmidrule(lr){4-4}
    Language & F1 (thres.) & F1 (thres.) & F1 (thres.)\\
    \midrule
         
        All & 0.63$\pm$0.05& 0.60$\pm$0.05& \textbf{0.76}$\pm$0.05\\
             & (0.88)& (0.82)& (0.82)\\
        Bengali & 0.63$\pm$0.09& 0.65$\pm$0.11& \textbf{0.67}$\pm$0.12\\
             & (0.87)& (0.72)& (0.79)\\
        English & 0.90$\pm$0.09& 0.81$\pm$0.12& \textbf{0.95}$\pm$0.08\\
             & (0.85)& (0.77)& (0.78)\\
        Hindi & \textbf{0.82}$\pm$0.09& 0.64$\pm$0.11& \textbf{0.82}$\pm$0.09\\
             & (0.88)& (0.77)& (0.82)\\
        Malayalam & 0.52$\pm$0.21& 0.62$\pm$0.17& \textbf{0.76}$\pm$0.16\\
             & (0.92)& (0.85)& (0.85)\\
        Tamil & 0.42$\pm$0.16& 0.54$\pm$0.18& \textbf{0.68}$\pm$0.13\\
             & (0.89)& (0.84)& (0.82)\\

    \bottomrule
    \end{tabular}
    
\end{table*}

\begin{table*}[h!]
    \centering
    \caption{Label distribution for the claim matching dataset: VS is very similar, SS is somewhat similar, SD is somewhat dissimilar and VD is very dissimilar. NM refers to ``no majority'' meaning there wasn't consensus among annotators.}
    \label{tbl:claim-matching-distribution}
    \begin{tabular}{lrrrrrrrrrr}
    \toprule
    & \multicolumn{2}{c}{VS} & \multicolumn{2}{c}{SS} & \multicolumn{2}{c}{SD} & \multicolumn{2}{c}{VD} & \multicolumn{2}{c}{NM}\\
    \cmidrule(lr){2-3}\cmidrule(lr){4-5}\cmidrule(lr){6-7}\cmidrule(lr){8-9}\cmidrule(lr){10-11}
    Language & \# & (\%) & \# & (\%) & \# & (\%) & \# & (\%) & \# & (\%) \\
    \midrule
         All & 261 & (11\%) & 121 & (5\%) & 115 & (5\%) & 1,417 & (61\%) & 429 & (18\%)\\
         Bengali & 38 & (6\%) & 62 & (10\%) & 26 & (4\%) & 225 & (35\%) & 293 & (45\%)\\
         English & 64 & (16\%) & 10 & (3\%) & 21 & (5\%) & 300 & (75\%) & 3 & (1\%) \\
         Hindi & 84 & (21\%) & 29 & (7\%) & 10 & (3\%) & 259 & (65\%) & 17 & (4\%) \\
         Malayalam & 42 & (7\%) & 9 & (2\%) & 51 & (8\%) & 474 & (78\%) & 28 & (5\%) \\
         Tamil & 33 & (11\%) & 11 & (4\%) & 7 & (2) & 159 & (53\%) & 88 & (30\%) \\
    \bottomrule
    \end{tabular}
\end{table*}

\end{document}